\documentclass[runningheads]{llncs}
\usepackage{graphicx}
\usepackage{comment}
\usepackage{amsmath,amssymb} 
\usepackage{color}
\usepackage{epsfig}
\usepackage{graphicx}
\usepackage{mathtools}
\usepackage{adjustbox}
\usepackage{gensymb}
\usepackage{indentfirst}
\usepackage{multirow}
\usepackage{mathrsfs}
\usepackage{booktabs}
\usepackage{array}
\usepackage{colortbl}
\usepackage{subfig}
\usepackage{lipsum}
\usepackage{contour}
\usepackage[pagebackref=true,
breaklinks=true,letterpaper=true,colorlinks,
citecolor=blue,
bookmarks=false]{hyperref}

\begin{document}
\pagestyle{headings}
\mainmatter
\def\ECCVSubNumber{2503}  

\title{Content-Aware Unsupervised Deep \\Homography Estimation} 

\titlerunning{Content-Aware Unsupervised Deep Homography Estimation}
\authorrunning{J. Zhang, C. Wang, S. Liu, L. Jia, N. Ye, J. Wang, J. Zhou and J. Sun}

\makeatletter
\renewcommand*{\@fnsymbol}[1]{\ensuremath{\ifcase#1\or \ddagger\or \dagger\or *\or
    \mathsection\or \mathparagraph\or \|\or **\or \dagger\dagger
    \or \ddagger\ddagger \else\@ctrerr\fi}}
\makeatother


\author{
Jirong Zhang\inst{1,2,*} \and
Chuan Wang\inst{2,*}\and
Shuaicheng Liu\inst{1,2,\dag} \and
Lanpeng Jia\inst{2}\and \\
Nianjin Ye\inst{2}\and
Jue Wang\inst{2}\and
Ji Zhou\inst{1}\and
Jian Sun\inst{2}
}
\institute{
University of Electronic Science and Technology of China \\
\email{\{zhangjirong@std., liushuaicheng@, jzhou233@\}uestc.edu.cn}
\and
Megvii Technology \\
\email{\{wangchuan, jialanpeng, yenianjin, wangjue, sunjian\}@megvii.com} \\
\url{https://github.com/JirongZhang/DeepHomography} \\
\inst{*}Joint First Authors,~~~\inst{\dag}Corresponding Author \\
Accepted by ECCV 2020\thanks{This is the arXiv version of our ECCV 2020 paper (Oral, Top 2\%, with 3/3 Strong Accepts), with more details revealed.}
}

\maketitle

\newcommand{\secname}{Sec.}

\begin{abstract}
Homography estimation is a basic image alignment method in many applications. It is usually conducted by extracting and matching sparse feature points, which are error-prone in low-light and low-texture images. On the other hand, previous deep homography approaches use either synthetic images for supervised learning or aerial images for unsupervised learning, both ignoring the importance of handling depth disparities and moving objects in real world applications. To overcome these problems, in this work we propose an unsupervised deep homography method with a new architecture design. In the spirit of the RANSAC procedure in traditional methods, we specifically learn an outlier mask to only select reliable regions for homography estimation. We calculate loss with respect to our learned deep features instead of directly comparing image content as did previously. To achieve the unsupervised training, we also formulate a novel triplet loss customized for our network. We verify our method by conducting comprehensive comparisons on a new dataset that covers a wide range of scenes with varying degrees of difficulties for the task. Experimental results reveal that our method outperforms the state-of-the-art including deep solutions and feature-based solutions.

\end{abstract}

\section{Introduction}\label{sec:intro}
Homography can align images taken from different perspectives if they approximately undergo a rotational motion or the scene is close to a planar surface~\cite{hartley2003multiple}. For scenes that satisfy the constraints, a homography can align them directly. For scenes that violate the constraints, e.g., a scene that consists of multiple planes or contains moving objects, homography usually serves as an initial alignment model before more advanced models such as mesh flow~\cite{liu2016meshflow} and optical flow~\cite{ilg2017flownet}. Most of the time, such a pre-alignment is crucial for the final quality. As a result, the homography has been widely applied in vision tasks such as multi-frame HDR imaging~\cite{gelfand2010multi}, multi-frame image super resolution~\cite{wronski2019handheld}, burst image denoising~\cite{liu2014fast}, video stabilization~\cite{liu2013bundled}, image/video stitching~\cite{zaragoza2013projective,guo2016joint}, SLAM~\cite{mur2015orb,zou2012coslam}, augmented reality~\cite{simon2000markerless} and camera calibration~\cite{zhang2000flexible}.

\begin{figure}[t]
  \centering
  \includegraphics[width=0.99\linewidth]{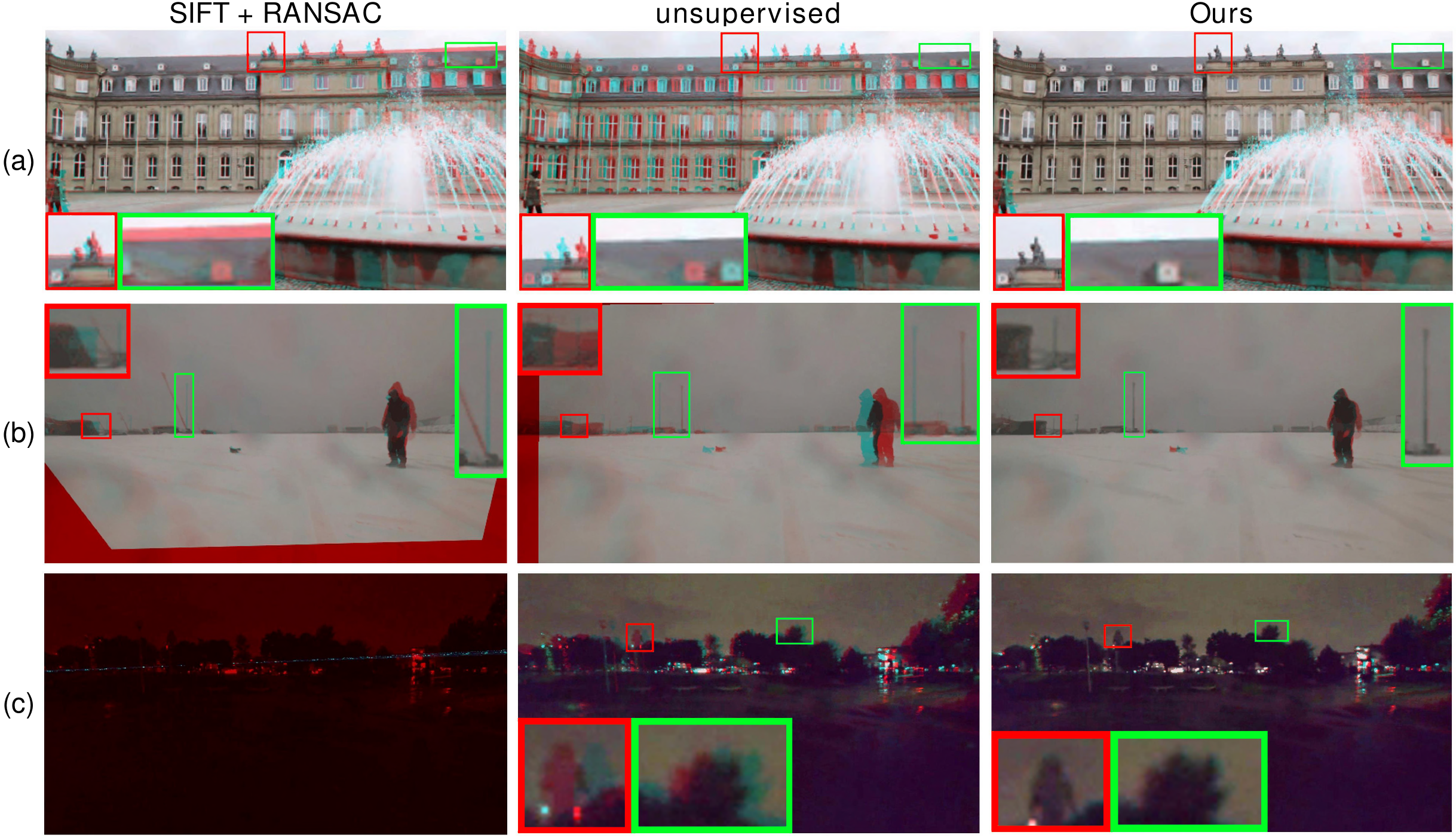}\\
  \caption{Our deep homography estimation on challenging cases, compared with one traditional feature-based, i.e. SIFT~\cite{lowe2004distinctive} + RANSAC and one unsupervised DNN-based method~\cite{nguyen2018unsupervised}. (a) An example with dominate moving foreground. (b) A low texture example. (c) A low light example. We mix the blue and green channels of the warped image and the red channel of the target image to obtain the visualization results as above, where the misaligned pixels appear as red or green ghosts. The same visualization method is applied for the rest of this paper.}\label{fig:teaser} 
\end{figure}

Homography estimation by traditional approaches generally requires matched image feature points such as SIFT~\cite{lowe2004distinctive}. Specifically, after a set of feature correspondences are obtained, a homography matrix is estimated by Direct Linear Transformation (DLT)~\cite{hartley2003multiple} with RANSAC outlier rejection~\cite{fischler1981random}. Feature-based methods commonly could achieve good performance while they highly rely on the quality of image features. Estimation could be inaccurate due to insufficient number of matched points or poor distribution of the features, which is a common case due to the existence of textureless regions (e.g., blue sky and white wall), repetitive patterns or illumination variations. Moreover, the rejection of outlier points, e.g., point matches that located on the non-dominate planes or dynamic objects, is also important for high quality results. Consequently, feature-based homography estimation is usually a challenging task for these non-regular scenes.

Due to the development of deep neural networks (DNN) in recent years, DNN-based solutions to homography estimation are gradually proposed such as supervised~\cite{detone2016deep} and unsupervised~\cite{nguyen2018unsupervised} ones. For the former solution, it requires homography as ground truth (GT) to supervise the training, so that only synthetic target images warped by the GT homography could be generated. Although the synthetic image pairs can be produced in arbitrary scale, they are far from real cases because real depth disparities are unavailable in the training data. As such, this method suffers from bad generalization to real images. To tackle this issue, Nguyen~\emph{et al.} proposed the latter unsupervised solution~\cite{nguyen2018unsupervised}, which minimizes the photometric loss on real image pairs. However, this method has two main problems. One is that the loss calculated with respect to image intensity is less effective than that in the feature space, and the loss is calculated uniformly in the entire image ignoring the RANSAC-like process. As a result, this method cannot exclude the moving or non-planar objects to contribute the final loss, so as to potentially decrease the estimation accuracy. To avoid the above phenomenons, Nguyen~\emph{et al.}~\cite{nguyen2018unsupervised} has to work on aerial images that are far away from the camera to minimize the influence of depth variations of parallax.

To tackle the aforementioned issues, we propose an unsupervised solution to homography estimation by a new architecture with content-awareness learning. It is designed specially for image pairs with a \textbf{small baseline}, as this case is commonly applicable for consecutive video frames, burst image capturing or photos captured by a dual-camera cellphone. In particular, to robustly optimize a homography, our network implicitly learns a deep feature for alignment and a content-aware mask to reject outlier regions simultaneously. The learned feature is used for loss calculation instead of using photometric loss as in~\cite{detone2016deep}, {\color{black}and learning a content-aware mask makes the network concentrate on the important and registrable regions.} We further formulate a novel triplet loss to optimize the network so that the unsupervised learning could be achieved. Experimental results demonstrate the effectiveness of all the newly involved techniques for our network, and qualitative and quantitative evaluations also show that our network outperforms the state-of-the-art as shown in Figs.~\ref{fig:teaser},~\ref{fig:sota} and~\ref{fig:feature_failure}. We also introduce a comprehensive image pair dataset, which contains 5 categories of scenes as well as human-labeled GT point correspondences for quantitative evaluation of its validation set (Fig.~\ref{fig:dataset}). To summarize, our main contributions are:
\begin{itemize}
  \item A novel network structure that enables content-aware robust homography estimation from two images with small baseline.
  \item A triplet loss designed for unsupervised training, so that an optimal homography matrix could be produced as an output, together with a deep feature map for alignment and a mask highlighting the alignment inliers being implicitly learned as intermediate results.
  \item A comprehensive dataset covers various scenes for unsupervised training of image alignment models, including but not limited to homography, mesh warps or optical flow.
\end{itemize}

\section{Related Work}\label{sec:related}

\subsubsection{Traditional homography.}
A homography is a $3 \times 3$ matrix which compensates plane motions between two images. It consists of $8$ degree of freedom (DOF), with each 2 for scale, translation, rotation and perspective~\cite{hartley2003multiple} respectively. To solve a homography, traditional approaches often detect and match image features, {\color{black}such as SIFT~\cite{lowe2004distinctive}, SURF~\cite{bay2006surf}, ORB~\cite{rublee2011orb}, LPM~\cite{ma2019locality}, GMS~\cite{bian2017gms}, SOSNet~\cite{tian2019sosnet}, LIFT~\cite{yi2016lift} and OAN~\cite{zhang2019learning}.} Two sets of correspondences were established between two images, following which robust estimation is adopted, such as the classic RANSAC~\cite{fischler1981random}, IRLS~\cite{holland1977robust} and MAGSAC~\cite{barath2019magsac}, for the outlier rejection during the model estimation. A homogrpahy can also be solved directly without image features. The direct methods, such as seminal Lucas-Kanade algorithm~\cite{lucas1981iterative}, calculates sum of squared differences (SSD) between two images. The differences guide the shift of the images, yielding homography updates. A random initialized homography is optimized in this way iteratively~\cite{baker2004lucas}. Moreover, the SSD can be replaced with enhanced correlation coefficient (ECC) for the robustness~\cite{evangelidis2008parametric}.

\subsubsection{Deep homography.}
Following the success of various deep image alignment methods such as optical flow~\cite{weinzaepfel2013deepflow,ilg2017flownet}, dense matching~\cite{revaud2016deepmatching}, learned descriptors~\cite{tian2019sosnet} and deep features~\cite{altwaijry2016learning}, a deep homography solution was first proposed by~\cite{detone2016deep} in 2016. The network takes source and target images as input and produces 4 corner displacement vectors of source image, so as to yield the homography. It used GT homography to supervise the training. However, the training images with GT homography is generated without depth disparity. To overcome such issue, Nguyen~\emph{et al.}~\cite{nguyen2018unsupervised} proposed an unsupervised approach that computed photometric loss between two images and adopted Spatial Transform Network (STN)~\cite{jaderberg2015spatial} for image warping. {\color{black}However, they calculated loss directly on the intensity and uniformly on the image plane. In contrast, we learn a content-aware mask. Notebaly, predicting mask for effective estimation has been attempted in other tasks, such as monocular depth estimation~\cite{zhou2017unsupervised,godard2019digging}. Here, it is introduced for the unsupervised homography learning.}


\begin{figure*}[t!]
  \centering
  \subfloat[][Network structure]{
  \includegraphics[width=0.62\linewidth]{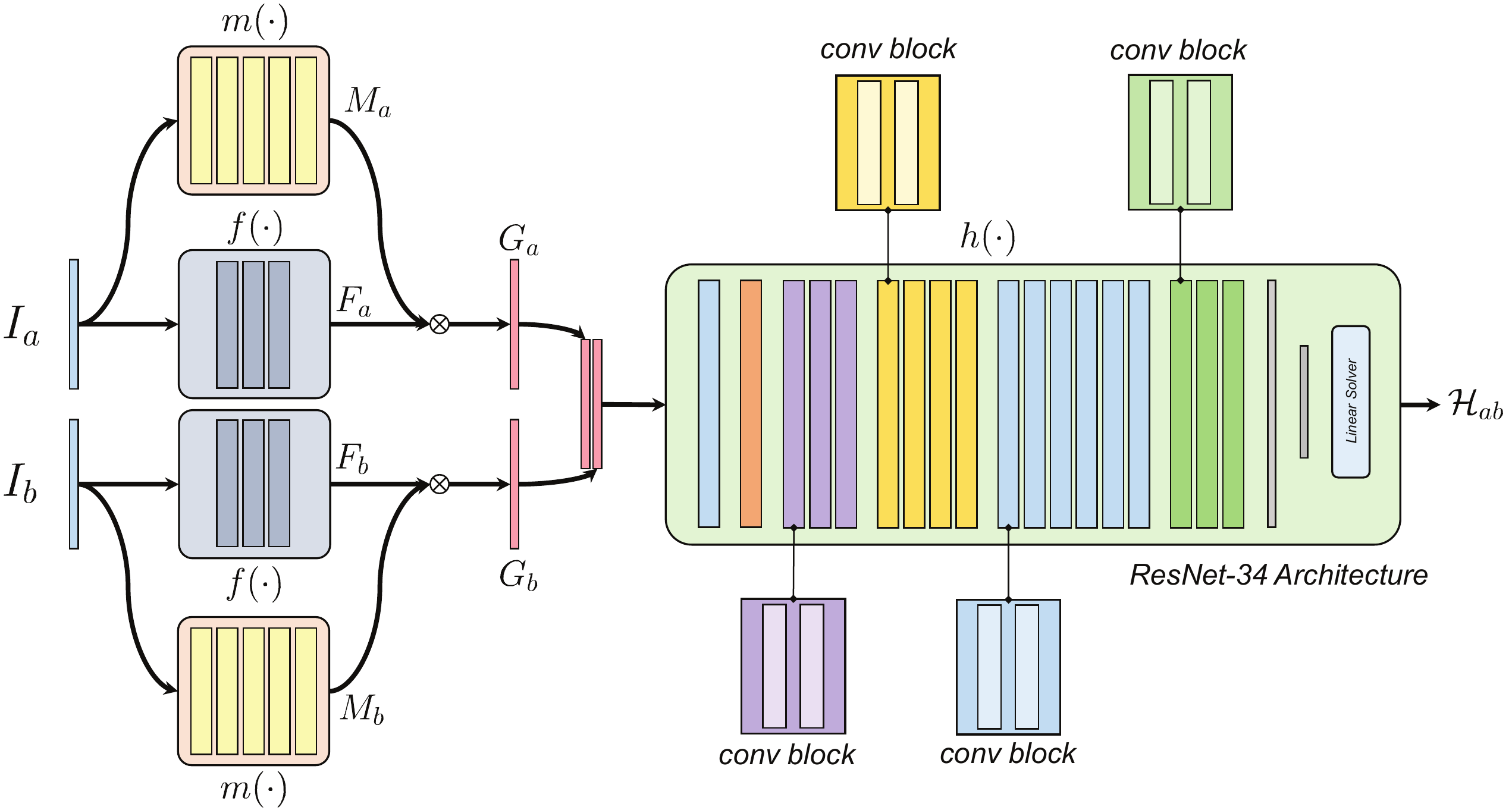}}
  \hfill
  \subfloat[][Triplet loss]{
  \includegraphics[width=0.35\linewidth]{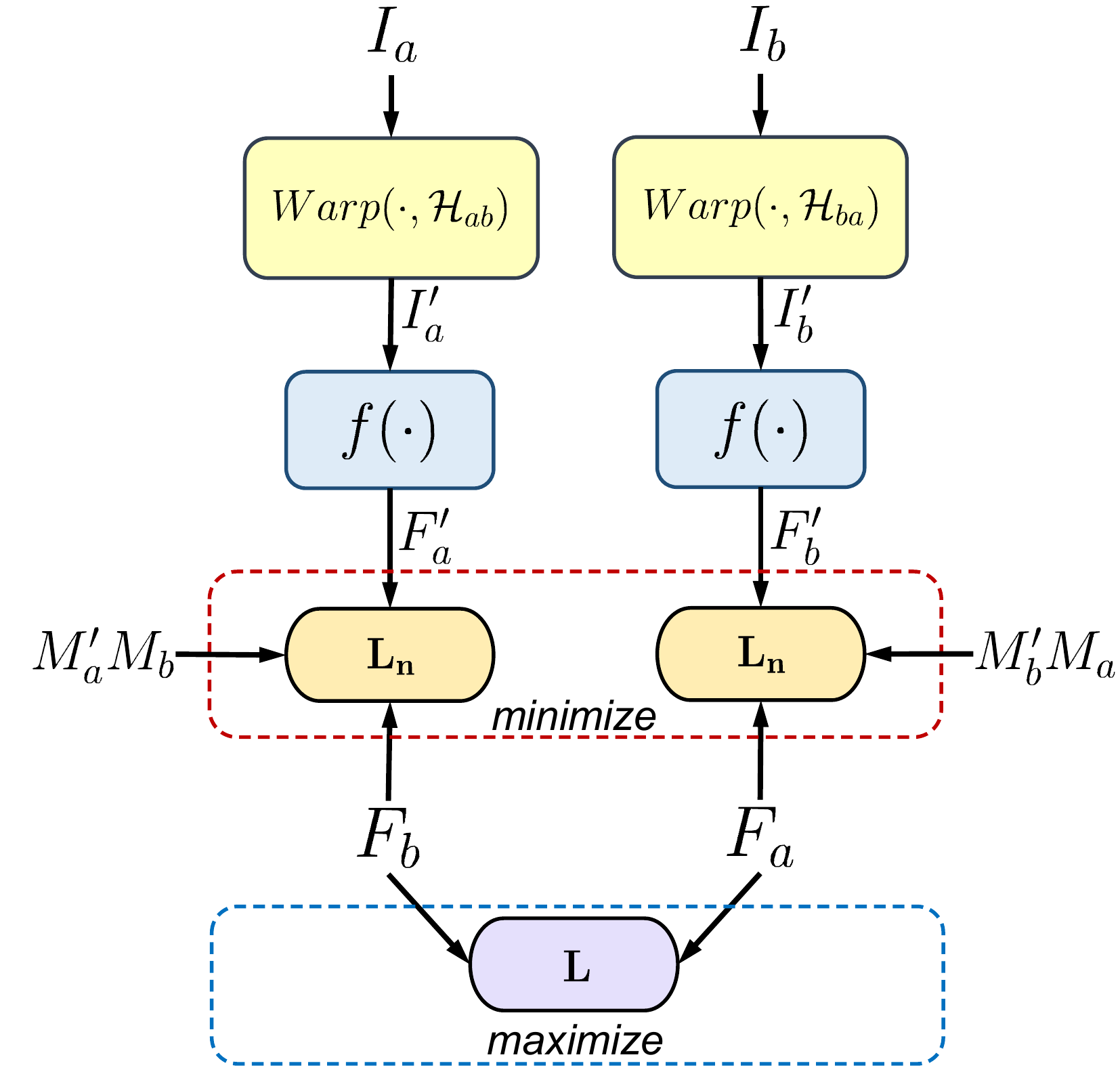}}
  \caption{The overall structure of our deep homography estimation network (a) and the triplet loss we design to train the network (b). In (a), two input patches $I_a$ and $I_b$ are fed into two branches consisting of feature extractor $f(\cdot)$ and mask predictor $m(\cdot)$ respectively, generating features $F_a, F_b$ and masks $M_a, M_b$. Then the features and masks are fed into a homography estimator to produce 8 values of the homography matrix $\mathcal{H}_{ab}$. In $h(\cdot)$, convolution blocks in various colors differ in the number of channels (detailed in Table~\ref{tab:net-layers}). To train the network in (a), we design a triplet loss composed of $\mathbf{L_n},\mathbf{L}$ as defined in Eq.~\ref{eq:l1-warp-Ia-Ib},~\ref{eq:l1-Ia-Ib} and~\ref{eq:tripleloss}.
  }\label{fig:network-structure}
\end{figure*}
\subsubsection{Image stitching.}
Image stitching methods~\cite{zaragoza2013projective,lin2017direct} are traditional methods that focus on stitching images under large baselines~\cite{zhang2014parallax} for the purpose of constructing the panorama~\cite{brown2003recognising}. The stitched images were often captured with dramatic viewpoint differences. In this work, we focus on images with small baselines for the purpose of multi-frame applications. 


\section{Algorithm}\label{sec:algo}

\subsection{Network Structure}
Our method is built upon convolutional neural networks. It takes two grayscale image patches $I_a$ and $I_b$ as input, and produces a homography matrix $\mathcal{H}_{ab}$ from $I_a$ to $I_b$ as output. The entire structure could be divided into three modules: a feature extractor $f(\cdot)$, a mask predictor $m(\cdot)$ and a homography estimator $h(\cdot)$. $f(\cdot)$ and $m(\cdot)$ are fully convolutional networks which accepts input of arbitrary sizes, and the $h(\cdot)$ utilizes a backbone of ResNet-34~\cite{he2016deep} and produces 8 values. \figurename~\ref{fig:network-structure}(a) illustrates the network structure. 

\subsubsection{Feature extractor.}
Unlike previous DNN based methods that directly utilizes the pixel intensity values as the feature, here our network automatically learns a deep feature from the input for robust feature alignment. To this end, we build a fully convolutional network (FCN) that takes an input of size $H\times W\times 1$, and produces a feature map of size $H\times W \times C$. For inputs $I_a$ and $I_b$, the feature extractor shares weights and produces feature maps $F_a$ and $F_b$, i.e.
\begin{align}
F_\beta = f(I_\beta), ~~\beta \in \{a, b\}
\end{align}

The learned feature is more robust than pixel intensity when applied to loss calculation. Especially for the images with luminance variations, the learned feature is pretty robust when compared to the pixel intensity. See \secname~\ref{subsec:ablation-content-aware-mask} and Fig.~\ref{fig:ablation-feat-ext} for a detailed verification of the effectiveness of this module. 

\begin{table}[t!]
\centering
\resizebox{0.4\linewidth}{!}
{
    \begin{tabular}{
    >{\centering\arraybackslash}p{2.2cm}
    >{\centering\arraybackslash}p{1.3cm}
    >{\centering\arraybackslash}p{1.3cm}
    >{\centering\arraybackslash}p{1.3cm}}
    \multicolumn{4}{c}{(a) Feature extractor $f(\cdot)$} \\
    \toprule
    Layer No. & 1     & 2     & 3 \\
    Type  & \textit{conv} & \textit{conv} & \textit{conv} \\ \midrule
    Kernel & 3     & 3     & 3 \\
    Stride & 1     & 1     & 1 \\
    Channel & 4     & 8     & 1 \\
    \bottomrule
    \end{tabular}
}\quad
\resizebox{0.57\linewidth}{!}
{
    \begin{tabular}{
    >{\centering\arraybackslash}p{2.2cm}
    >{\centering\arraybackslash}p{1.3cm}
    >{\centering\arraybackslash}p{1.3cm}
    >{\centering\arraybackslash}p{1.3cm}
    >{\centering\arraybackslash}p{1.3cm}
    >{\centering\arraybackslash}p{1.3cm}}
    \multicolumn{6}{c}{(b) Mask predictor $m(\cdot)$} \\ \toprule
    Layer No. & 1     & 2     & 3     & 4     & 5 \\
    Type  & \textit{conv} & \textit{conv} & \textit{conv} & \textit{conv} & \textit{conv} \\ \midrule
    Kernel & 3     & 3     & 3     & 3     & 3 \\
    Stride & 1     & 1     & 1     & 1     & 1 \\
    Channel & 4     & 8     & 16    & 32    & 1 \\ \bottomrule
    \end{tabular}
}
\resizebox{0.99\linewidth}{!}
{
\begin{tabular}{
>{\centering\arraybackslash}p{2.2cm}
>{\centering\arraybackslash}p{1.3cm}
>{\centering\arraybackslash}p{1.3cm}
>{\centering\arraybackslash}p{1.3cm}
>{\centering\arraybackslash}p{1.3cm}
>{\centering\arraybackslash}p{1.3cm}
>{\centering\arraybackslash}p{1.3cm}
>{\centering\arraybackslash}p{1.3cm}
>{\centering\arraybackslash}p{1.3cm}
>{\centering\arraybackslash}p{1.3cm}
>{\centering\arraybackslash}p{1.3cm}
>{\centering\arraybackslash}p{0.9cm}
}
\multicolumn{12}{c}{\vspace{-2mm}} \\
\multicolumn{12}{c}{(c) Homography estimator $h(\cdot)$} \\
\toprule
Layer No. & 1    & 2        & 3 $\sim$ 8 & 9    & 10 $\sim$ 16 & 17   & 18 $\sim$ 28 & 29   & 30 $\sim$ 34 & 35                  & 36 \\
Type      & \textit{conv} & \textit{pool} & \textit{conv}     & \textit{conv} & \textit{conv}       & \textit{conv} & \textit{conv}       & \textit{conv} & \textit{conv}       & \textit{pool} & \textit{fc} \\ \midrule
Kernel    & 7    & 3        & 3        & 3    & 3          & 3    & 3          & 3    & 3          & -                   & -  \\
Stride    & 2    & 2        & 1        & 2    & 1          & 2    & 1          & 2    & 1          & 1                   & -  \\
Channel   & 64   & -        & 64       & 128  & 128        & 256  & 256        & 512  & 512        & -                   & 8  \\
\bottomrule
\multicolumn{12}{c}{\vspace{-2mm}}
\end{tabular}
}
\caption{Layer configurations of feature extractor (a), mask predictor (b) and homography estimator (c).
In (c), Layer 2 and 35 are max pool and global average pool separately.}\label{tab:net-layers}
\end{table}

\subsubsection{Mask predictor.}
In non-planar scenes, especially those including moving objects, there exists no single homography that can align the two views. In traditional algorithm, RANSAC is widely applied to find the inliers for homography estimation, so as to solve the most approximate matrix for the scene alignment. Following the similar idea, we build a sub-network to automatically learn the positions of inliers. Specifically, a sub-network $m(\cdot)$ learns to produce an inlier probability map or mask, highlighting the content in the feature maps that contribute much for the homography estimation. The size of the mask is the same as the size of the feature maps $F_a$ and $F_b$. With the masks, we further weight the features extracted by $f$ before feeding them to the homography estimator, obtaining two weighted feature maps $G_a$ and $G_b$ as,
\begin{align}
M_\beta = m(I_\beta), ~~G_\beta = F_\beta M_\beta, \quad \beta \in \{a,b\}
\label{eq:GaGb}
\end{align}
As introduced later, the mask learned as above actually play two roles in the network, one works as an attention map, and the other works as a outlier rejecter. See the details in \secname~\ref{subsec:triplet-loss},~\ref{subsec:ablation-content-aware-mask} and Fig.~\ref{fig:mask} for more discussion.

\subsubsection{Homography estimator.}
Given the weighted feature maps $G_a$ and $G_b$, we concatenate them to build a feature map $[G_a, G_b]$ of size $H\times W\times 2C$. Then it is fed to the homography estimator network and four 2D offset vectors (8 values) are produced. With the $4$ offset vectors, it is straight-forward to obtain the homography matrix $\mathcal{H}_{ab}$ with 8 DOF by solving a linear system. We use $h(\cdot)$ to represent the whole process, i.e.
\begin{align}
\mathcal{H}_{ab} = h([G_a, G_b])
\end{align}
The backbone of $h(\cdot)$ follows a ResNet-34 structure. It contains 34 layers of strided convolutions followed by a global average pooling layer, which generates fixed size (8 in our case) of feature vectors regardless of the input feature dimensions. Please refer to Table~\ref{tab:net-layers} for more details. 

\subsection{Triplet Loss for Robust Homography Estimation}\label{subsec:triplet-loss}
With the homography matrix $\mathcal{H}_{ab}$ estimated, we warp image $I_a$ to $I'_a$ and then further extracts its feature map as $F'_a$. Intuitively, if the homography matrix $\mathcal{H}_{ab}$ is accurate enough, $F'_a$ should be well aligned with $F_b$, causing a low $l_1$ loss between them. Considering in real scenes, a single homography matrix normally cannot satisfy the transformation between the two views, we also normalize the $l_1$ loss by $M'_a$ and $M_b$. Here $M'_a$ is the warped version of $M_a$. So the loss between the warped $I_a$ and $I_b$ is as follows,
\begin{align}
\mathbf{L_n}(I'_a, I_b) = \frac{\sum_i M'_aM_b\cdot||F'_a - F_b||_1}{\sum_i M'_aM_b} \label{eq:l1-warp-Ia-Ib}
\end{align}
where $F'_a = f(I'_a)$ and $I'_a = Warp(I_a, \mathcal{H}_{ab})$. Index $i$ indicates pixel locations in the masks and feature maps. STN~\cite{jaderberg2015spatial} is used to achieve the warping operation.

Directly minimizing Eq.~\ref{eq:l1-warp-Ia-Ib} may easily cause trivial solutions, where the feature extractor only produces all zero maps, i.e. $F'_a = F_b = 0$. In this case, the features learned indeed describe the fact that $I'_a$ and $I_b$ are ``well aligned'', but it fails to reflect the fact that the original images $I_a$ and $I_b$ are mis-aligned. To this end, we involve another loss between $F_a$ and $F_b$, i.e.
\begin{align}
\mathbf{L}(I_a, I_b) = ||F_a - F_b||_1 \label{eq:l1-Ia-Ib}
\end{align}
and further maximize it when minimizing Eq.~\ref{eq:l1-warp-Ia-Ib}. This strategy avoids the trivial all-zero solutions, and enables the network to learn a discriminative feature map.

\begin{figure*}[t]
  \centering
  \includegraphics[width=0.98\linewidth]{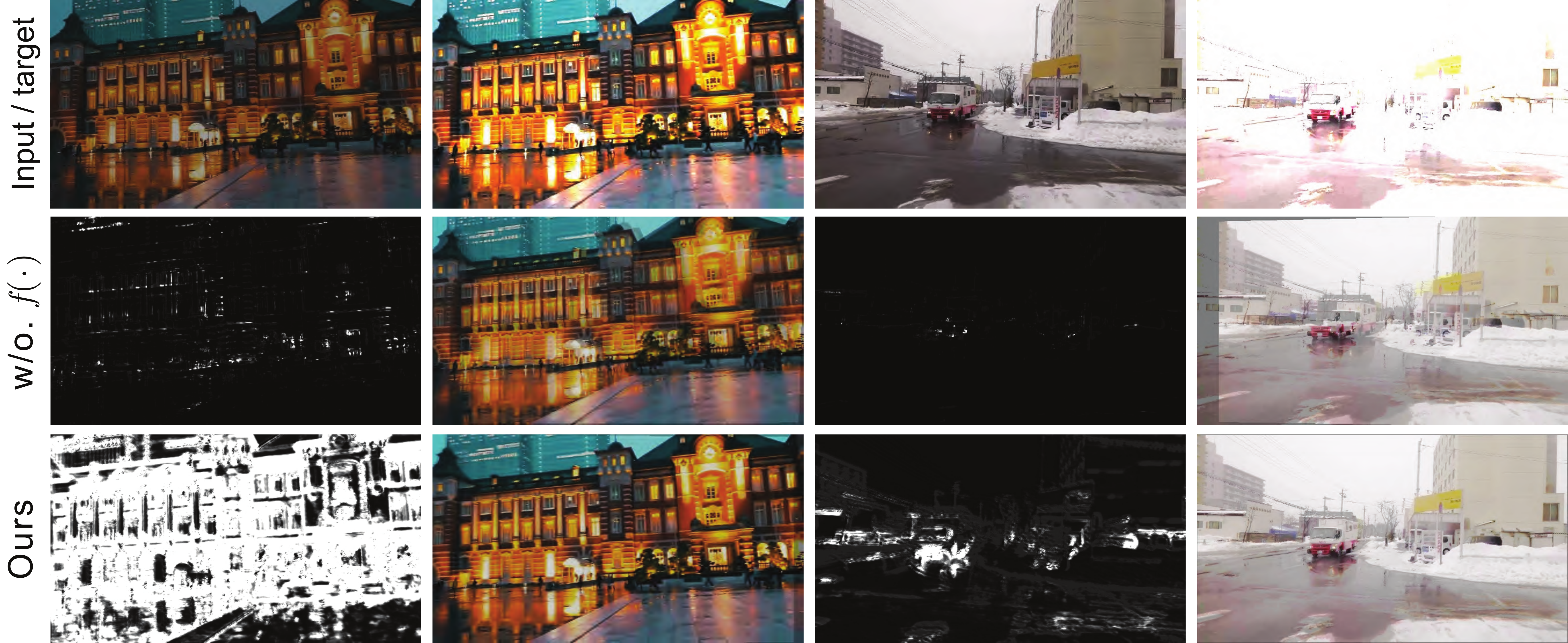}
  \caption{Ablation study on the effectiveness of our feature extractor, demonstrated by examples with illuminance change, displayed separately in the left and right two columns. For each example, the input and target GT images are in Row 1, followed by the results by disabling the feature extractor $f(\cdot)$ (Row 2) and by ours (Row 3), including the learned masks and the aligned results in odd and even columns. As seen, our results are obviously stable for such a case.}
  \label{fig:ablation-feat-ext}
\end{figure*}

\begin{figure*}[t]
  \centering
  \includegraphics[width=0.95\linewidth]{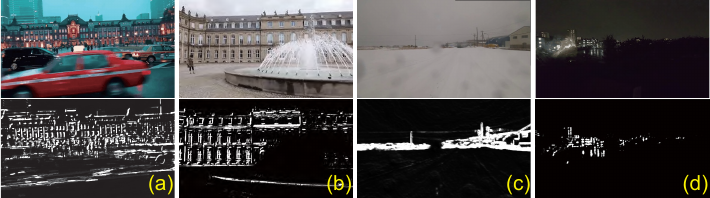}\\
  \makebox[\linewidth]{\rule{0.95\linewidth}{0.4pt}}
  \includegraphics[width=0.95\linewidth]{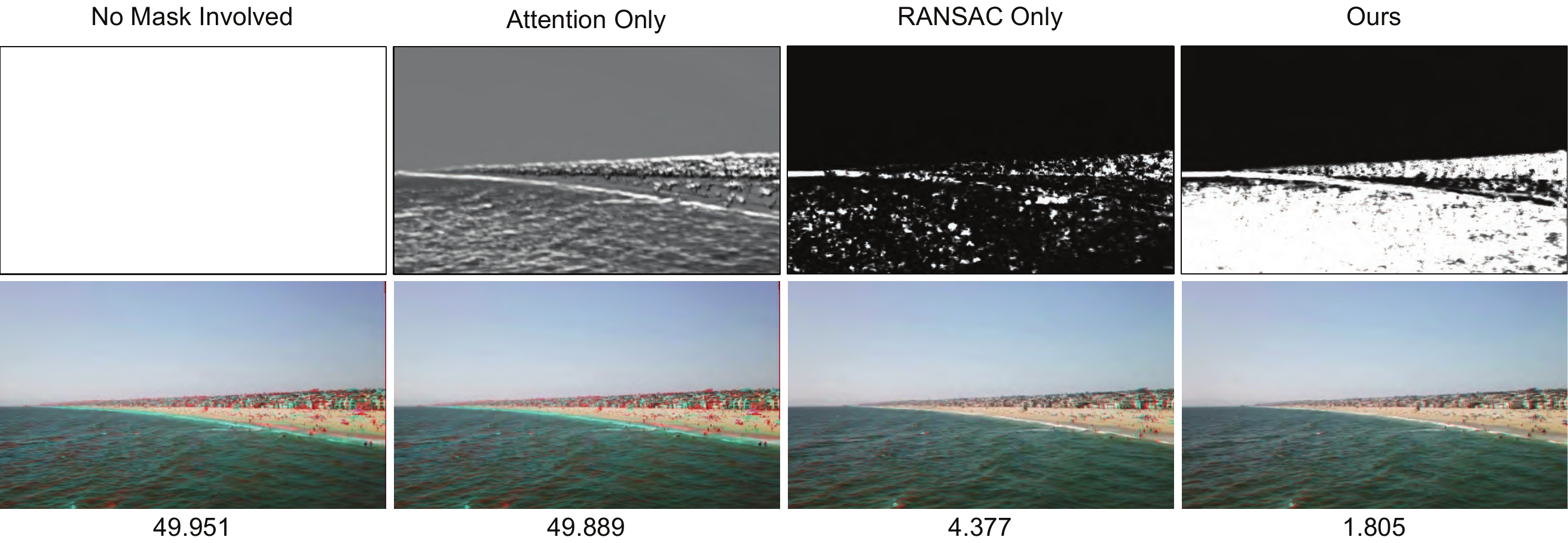}
  \caption{ Row 1 and 2: Our predicted masks for various of scenes. (a) and (b) contains large dynamic foreground. (c) contains few textures and (d) is an night example. Row 3 and 4: Ablation study on the content-aware mask. We disable both or either role of the mask for comparisons. Errors are shown at the bottom.}
  \label{fig:mask}
\end{figure*}

In practise, we swap the features of $I_a$ and $I_b$ and produce another homography matrix $\mathcal{H}_{ba}$. Following Eq.~\ref{eq:l1-warp-Ia-Ib}, we involve a loss $\mathbf{L_n}(I'_b,I_a)$ between the warped $I_b$ and $I_a$. We also add a constraint that enforces $\mathcal{H}_{ab}$ and $\mathcal{H}_{ba}$ to be inverse. So, the optimization procedure of the network is written as follows,
\begin{align}
\min_{m, f, h}  \mathbf{L_n}(I'_a, I_b) + \mathbf{L_n}(I'_b, I_a) - \lambda \mathbf{L}(I_a, I_b) 
 + \mu ||\mathcal{H}_{ab}\mathcal{H}_{ba} - \mathcal{I}||_2^2
\label{eq:tripleloss}
\end{align}
 where $\lambda$ and $\mu$ are balancing hyper-parameters, and $\mathcal{I}$ is a 3-order identity matrix. We set $\lambda = 2.0$ and $\mu = 0.01$ in our experiments. We show the loss formulations in Fig.~\ref{fig:network-structure}(b), and validate its effectiveness by an ablation study detailed in \secname~\ref{subsec:ablation-content-aware-mask}, which shows that it decreases the error at least 50\% in average.

\subsection{Unsupervised Content-Awareness Learning}\label{subsec:unsupervised-content-awareness-learning}
As mentioned above, our network contains a sub-network $m(\cdot)$ to predict an inlier probability mask. It is such designed that our network can be of content-awareness by the two-fold roles. First, we use the masks $M_a, M_b$ to explicitly weight the features $F_a, F_b$, so that only highlighted features could be fully fed into homography estimator $h(\cdot)$. The masks actually serve as attention maps for the feature maps. Second, they are also implicitly involved into the normalized loss Eq.~\ref{eq:l1-warp-Ia-Ib}, working as a weighting item.
By doing this, only those regions that are really fit for alignment would be taken into account. For those areas containing low texture or moving foreground, because they are non-distinguishable or misleading for alignment, they are naturally removed for homography estimation during optimizing the triplet loss as proposed. Such a content-awareness is achieved fully by an unsupervised learning scheme, without any GT mask data as supervision. To demonstrate the effectiveness of the mask as the two roles, we conduct an ablation study by disabling the effect of mask working as an attention map or as a loss weighting item. As seen in Table~\ref{tab:comp-all-methods}(c), the accuracy has a significant decrease when mask is removed in either case.

We also illustrate several examples in Fig.~\ref{fig:mask} to show the mask effectiveness. For example, in {\color{black}Fig.~\ref{fig:mask}(a)(b) where the scenes contain large dynamic foregrounds, our network successfully rejects moving objects, even if the movements are inapparent as the fountain in (b), or the objects occupy a large space as in (a).} These cases are very difficult for RANSAC to find robust inliers. {\color{black}Fig.~\ref{fig:mask}(c) is a low-textured example, in which the sky and snow ground occupies almost the entire image. It is challenging for traditional methods because not enough feature matches can be provided. Our predicted mask concentrates on the horizon for the alignment.} Last, Fig.~\ref{fig:mask}(d) is a low light example, where only visible areas contain weights as seen. We also illustrate an example to show the two effects by the mask as separate roles in the bottom 2 rows of Fig.~\ref{fig:mask}. Details about this ablation study are introduced later in \secname~\ref{subsec:ablation-content-aware-mask}. 

We adopt a two-stage strategy to train our network. Specifically, we first train the network by disabling the attention map role of the mask, i.e. $G_\beta = F_\beta,~\beta \in \{a,b\}$. After about $60k$ iterations, we finetune the network by involving the attention map role of the mask as Eq.~\ref{eq:GaGb}. We validate this training strategy by another ablation study detailed in \secname~\ref{subsec:ablation-training-strategy}, where we train the network totally from scratch. This two-stage training strategy reduces the error by 4.40\% in average, as shown in Row 10 of Table~\ref{tab:comp-all-methods}(c).
\section{Experimental Results}\label{sec:exp}

\begin{figure}[t]
  \centering
  \includegraphics[width=0.99\linewidth]{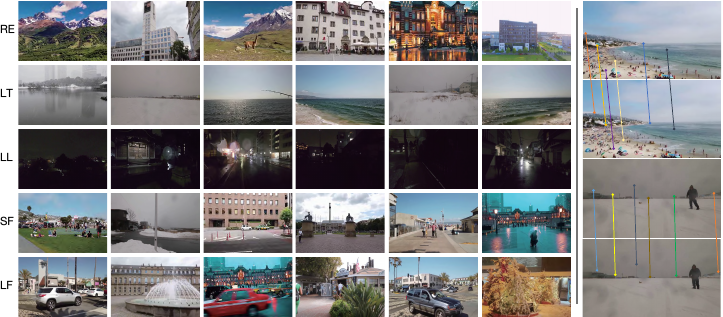}\\
  \caption{A glace of our dataset. For left 6 columns, from top to bottom are the 5 categories of the dataset. The rightmost column shows two examples of human labeled point correspondences for quantitative evaluation.}
  \label{fig:dataset}
\end{figure}%

\subsection{Dataset and Implementation Details}

We propose our dataset for comprehensive homography evaluation considering there lacks dedicated dataset for this task. Our dataset contains $5$ categories of totally $80k$ image pairs, including regular (\textbf{RE}), low-texture (\textbf{LT}), low-light (\textbf{LL}), small-foregrounds (\textbf{SF}), and large-foregrounds (\textbf{LF}) scenes, with each category $\approx16k$ image pairs, as shown in Fig.~\ref{fig:dataset}.
For the test data, $4.2k$ image pairs are randomly chosen from all categories. For each pair, we manually marked $6 \sim 8$ equally distributed matching points for the purpose of quantitative comparisons, as illustrated in the rightmost column of Fig.~\ref{fig:dataset}. {\color{black}The category partition is based on the understanding and property of traditional homography registration.} Experimental results demonstrate our method is robust over all categories as seen in Figs.~\ref{fig:teaser},~\ref{fig:sota},~\ref{fig:feature_failure} and the supplementary materials, which also contain a detailed introduction to each category. 

Our network is trained with $120k$ iterations by an Adam optimizer~\cite{kingma2014adam}, with parameters being set as $l_r = 1.0\times 10^{-4}$, $\beta_1 = 0.9$, $\beta_2 = 0.999$, $\varepsilon = 1.0\times 10^{-8}$. The batch size is $64$, and for every $12k$ iterations, the learning rate $l_r$ is reduced by 20\%. Each iteration costs about $1.2$s and it takes nearly $40$ hours to complete the entire training. The implementation is based on PyTorch and the network training is performed on $4$ NVIDIA RTX 2080 Ti. To augment the training data and avoid black boundaries appearing in the warped image, we randomly crop patches of size $315\times560$ from the original image to form $I_a$ and $I_b$. Code is available at \url{https://github.com/JirongZhang/DeepHomography}.
\subsection{Comparisons with Existing Methods}
\subsubsection{Qualitative comparison.}

\begin{figure*}[p]
  \centering
  \includegraphics[width=1.0\linewidth]{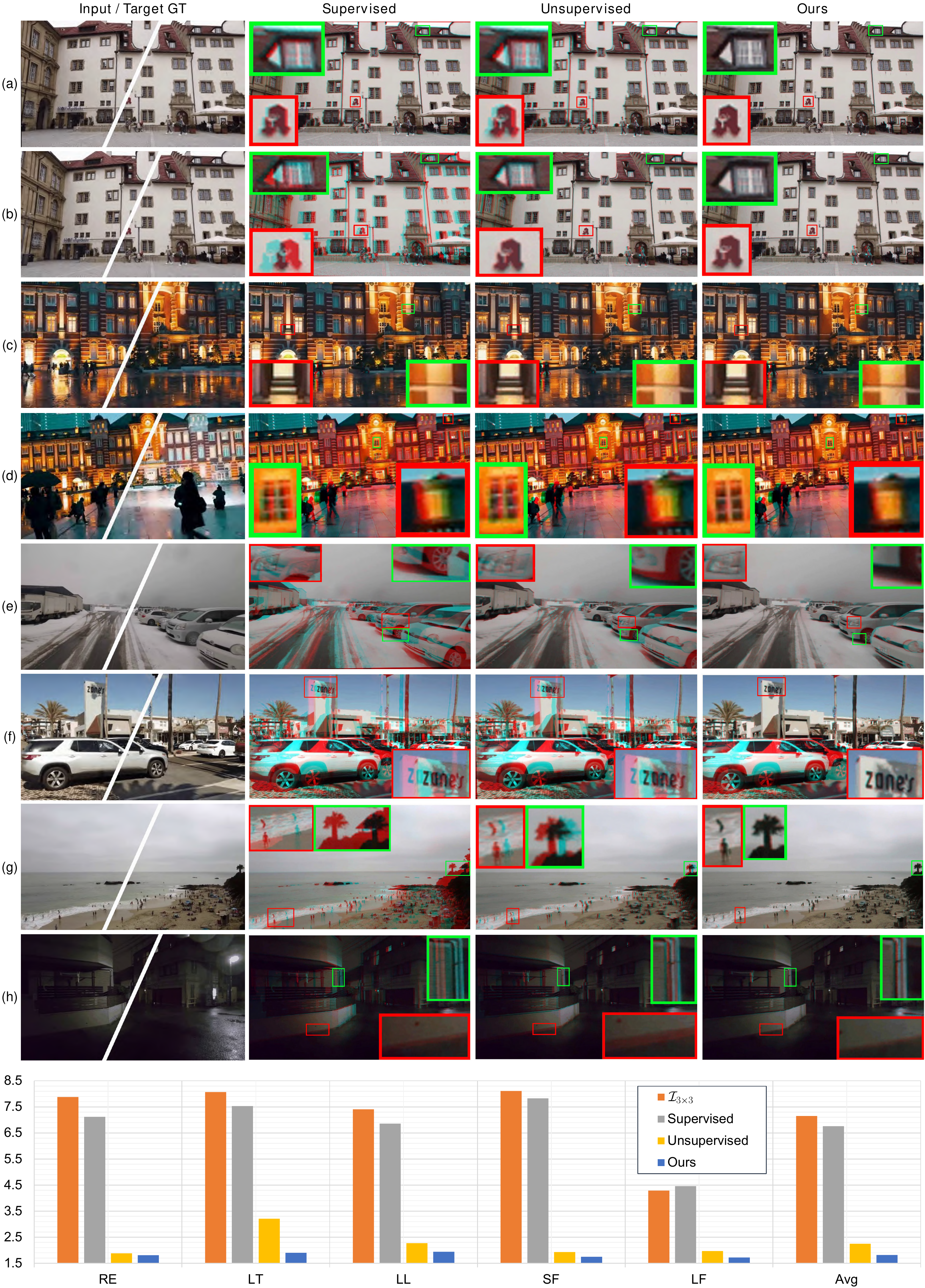}
  \caption{Comparison with existing DNN-based approaches. Column 1 shows the input and GT target images, columns 2 to 4 are results by the supervised~\cite{detone2016deep}, the unsupervised~\cite{nguyen2018unsupervised} and our method. The errors by all the DNN-based methods are displayed by a bar chart at the bottom.
  }\label{fig:sota}
\end{figure*}

We first compare our method with the existing two deep homography methods, the supervised~\cite{detone2016deep} and the unsupervised~\cite{nguyen2018unsupervised} approaches, as illustrated in Fig.~\ref{fig:sota}. Fig.~\ref{fig:sota}(a) shows an synthesized example with no disparities. In this case, the supervised solution~\cite{detone2016deep} performs well enough as ours. However, it fails in the case that real consecutive frames of the same footage are applied (Fig.~\ref{fig:sota}(b)), because it is unable to handle large disparities and moving objects of the scene. Fig.~\ref{fig:sota}(c) shows an example that contains a dominate planar building surface, where all methods work well. However, if the image pair involves illumination variation caused by camera flash, the unsupervised method~\cite{nguyen2018unsupervised} fails due to its alignment metric being pixel intensity value difference instead of semantic feature difference, as seen in Fig.~\ref{fig:sota}(d). 
Fig.~\ref{fig:sota}(e) and (f) contain near-range objects and two dominate planes with moving objects at corners respectively, and Fig.~\ref{fig:sota}(g) and (h) are low texture and low light examples separately. Similarly, in all of these scenarios, our method produces warped images with more pixels aligned, so as to obviously outperform the other two DNN-based methods.

\newcommand{\hltred}[1]{\textcolor{red}{\textbf{#1}}}
\newcommand{\hltblue}[1]{\textcolor{blue}{#1}}
\begin{table}[t!]
\centering
  \resizebox{1.0\linewidth}{!}{
    \begin{tabular}{
    r
    p{4.cm}
    >{\centering\arraybackslash}p{2.9cm}
    >{\centering\arraybackslash}p{2.9cm}
    >{\centering\arraybackslash}p{2.9cm}
    >{\centering\arraybackslash}p{2.9cm}
    >{\centering\arraybackslash}p{2.9cm}
    >{\centering\arraybackslash}p{2.9cm}}
    \multicolumn{8}{c}{\textbf{{\large{(a) Errors}}}} \\
    \toprule
    1) &     & RE    & LT    & LL    & SF    & LF    & Avg \\
    \midrule
    2) & $\mathcal{I}_{3\times3}$ & 7.88 (+360.82\%) & 8.07 (+215.23\%) & 7.41 (+252.86\%) & 8.11 (+360.80\%) & 4.29 (+142.37\%) & 7.15 (+245.41\%) \\ \midrule
    3) &Supervised~\cite{detone2016deep} & 7.12 (+316.37\%) & 7.53 (+194.14\%) & 6.86 (+226.67\%) & 7.83 (+344.89\%) & 4.46 (+151.98\%) & 6.76 (+226.57\%) \\
    4) &Unsupervised~\cite{nguyen2018unsupervised} & 1.88 (+9.94\%) & 3.21 (+25.39\%) & 2.27 (+8.10\%) & 1.93 (+9.66\%) & 1.97 (+11.30\%) & 2.25 (+8.70\%) \\
    \midrule
    5) &SIFT~\cite{lowe2004distinctive} + RANSAC~\cite{fischler1981random} & 1.72 (+0.58\%) & \textcolor[rgb]{ 0,  0,  1}{2.56 (+0.00\%)} & 4.97 (+136.67\%) & 1.82 (+3.41\%) & 1.84 (+3.95\%) & 2.58 (+24.64\%) \\
    6) &SIFT~\cite{lowe2004distinctive} + MAGSAC~\cite{barath2019magsac} & \textcolor[rgb]{ 1,  0,  0}{\textbf{1.71 (+0.00\%)}} & 3.15 (+23.05\%) & 4.91 (+133.81\%) & 1.88 (+6.82\%) & 1.79 (+1.13\%) & 3.20 (+54.59\%) \\
    7) &ORB~\cite{rublee2011orb} + RANSAC~\cite{fischler1981random} & 1.85 (+8.19\%) & 3.76 (+46.88\%) & 2.56 (+21.90\%) & 2.00 (+13.64\%) & 2.29 (+29.38\%) & 2.49 (+20.29\%) \\
    8) &ORB~\cite{rublee2011orb} + MAGSAC~\cite{barath2019magsac} & 2.02 (+18.13\%) & 5.18 (+102.34\%) & 2.78 (+32.38\%) & 1.92 (+9.09\%) & 2.25 (+27.12\%) & 2.83 (+36.71\%) \\
    9) &LIFT~\cite{yi2016lift} + RANSAC~\cite{fischler1981random} & 1.76 (+2.92\%) & 3.04 (+18.75\%) & 2.14 (+1.90\%) & 1.82 (+3.41\%) & 1.92 (+8.47\%) & 2.14 (+3.38\%) \\
    10) &LIFT~\cite{yi2016lift} + MAGSAC~\cite{barath2019magsac} & 1.73 (+1.17\%) & 2.92 (+14.06\%) & \textcolor[rgb]{ 0,  0,  1}{2.10 (+0.00\%)} & 1.79 (+1.70\%) & 1.79 (+1.13\%) & \textcolor[rgb]{ 0,  0,  1}{2.07 (+0.00\%)} \\
    11) &SOSNet~\cite{tian2019sosnet} + RANSAC~\cite{fischler1981random} & 1.72 (+0.58\%) & 3.70 (+44.53\%) & 4.58 (+118.09\%) & 1.84 (+4.54\%) & 1.83 (+3.39\%) & 2.73 (+31.88\%) \\
    12) &SOSNet~\cite{tian2019sosnet} + MAGSAC~\cite{barath2019magsac} & 1.73 (+1.17\%) & 5.14 (+100.78\%) & 4.39 (+109.05\%) & \textcolor[rgb]{ 0,  0,  1}{1.76 (+0.00\%)} & \textcolor[rgb]{ 0,  0,  1}{1.77 (+0.00\%)} & 2.99 (+44.44\%) \\
    \midrule
    13) &Ours  & \textbf{1.81 (+5.85\%)} & \textcolor[rgb]{ 1,  0,  0}{\textbf{1.90  (-25.78\%)}} & \textcolor[rgb]{ 1,  0,  0}{\textbf{1.94  (-7.62\%)}} & \textcolor[rgb]{ 1,  0,  0}{\textbf{1.75  (-0.57\%)}} & \textcolor[rgb]{ 1,  0,  0}{\textbf{1.72  (-2.82\%)}} & \textcolor[rgb]{ 1,  0,  0}{\textbf{1.82  (-12.08\%)}} \\
    \bottomrule
    \end{tabular}
    }\vspace{1mm}
  \resizebox{1.0\linewidth}{!}{
    \begin{tabular}{
    r
    p{4.cm}
    >{\centering\arraybackslash}p{2.9cm}
    >{\centering\arraybackslash}p{2.9cm}
    >{\centering\arraybackslash}p{2.9cm}
    >{\centering\arraybackslash}p{2.9cm}
    >{\centering\arraybackslash}p{2.9cm}
    >{\centering\arraybackslash}p{2.9cm}}
    \multicolumn{8}{c}{\textbf{{\large{(b) Robustness: Inlier Percentage When Matched Points Are within $3$ Pixels}}}} \\
    \toprule
    1) & & RE    & LT    & LL    & SF    & LF    & Avg \\ \midrule
    2) & $\mathcal{I}_{3\times3}$ & 12.75\% (-85.35\%) & 37.83\% (-54.13\%) & 36.68\% (-55.32\%) & 48.46\% (-42.43\%) & 64.30\% (-25.15\%) & 38.76\% (-53.79\%) \\ \midrule
    3) & Supervised~\cite{detone2016deep} & 16.17\% (-81.42\%) & 42.76\% (-48.16\%) & 40.73\% (-50.38\%) & 48.24\% (-42.69\%) & 61.29\% (-28.65\%) & 40.89\% (-51.25\%) \\
    4) & Unsupervised~\cite{nguyen2018unsupervised} & 85.57\% (-1.69\%) & 71.41\% (-13.42\%) & 79.45\% (-3.22\%) & 82.52\% (-1.96\%) & 83.65\% (-2.62\%) & 79.80\% (-4.86\%) \\ \midrule
    5) & SIFT~\cite{lowe2004distinctive}+RANSAC~\cite{fischler1981random} & 86.95\% (-0.10\%) & 81.98\% (-0.61\%) & 80.79\% (-1.58\%) & \hltblue{84.17\% (+0.00\%)} & 85.36\% (-0.63\%) & 83.77\% (-0.13\%) \\
    6) & SIFT~\cite{lowe2004distinctive}+MAGSAC~\cite{barath2019magsac} & 86.70\% (-0.39\%) & \hltblue{82.48\% (+0.00\%)} & 80.67\% (-1.73\%) & 83.69\% (-0.57\%) & \hltblue{85.90\% (+0.00\%)} & \hltblue{83.88\% (+0.00\%)} \\
    7) & ORB~\cite{rublee2011orb}+RANSAC~\cite{fischler1981random} & 85.31\% (-1.99\%) & 77.21\% (-6.39\%) & 81.44\% (-0.79\%) & 83.55\% (-0.74\%) & 79.70\% (-7.22\%) & 81.00\% (-3.43\%) \\
    8) & ORB~\cite{rublee2011orb}+MAGSAC~\cite{barath2019magsac} & 83.55\% (-4.01\%) & 75.15\% (-8.89\%) & 80.77\% (-1.61\%) & 81.25\% (-3.47\%) & 79.80\% (-7.10\%) & 79.70\% (-4.98\%) \\
    9) & LIFT~\cite{yi2016lift}+RANSAC~\cite{fischler1981random} & 86.50\% (-0.62\%) & 72.58\% (-12.00\%) & 80.89\% (-1.46\%) & 83.22\% (-1.13\%) & 83.42\% (-2.89\%) & 80.63\% (-3.87\%) \\
    10) & LIFT~\cite{yi2016lift}+MAGSAC~\cite{barath2019magsac} & \hltred{87.04\% (+0.00\%)} & 74.53\% (-9.64\%) & \hltblue{82.09\% (+0.00\%)} & 83.84\% (-0.39\%) & 85.61\% (-0.34\%) & 82.03\% (-2.21\%) \\
    11) & SOSNet~\cite{tian2019sosnet}+RANSAC~\cite{fischler1981random} & 87.03\% (-0.01\%) & 81.44\% (-1.26\%) & 80.69\% (-1.71\%) & 84.10\% (-0.08\%) & 85.48\% (-0.49\%) & 83.63\% (-0.30\%) \\
    12) & SOSNet~\cite{tian2019sosnet}+MAGSAC~\cite{barath2019magsac} & 86.93\% (-0.13\%) & 81.81\% (-0.81\%) & 80.63\% (-1.78\%) & 83.29\% (-1.05\%) & 85.84\% (-0.07\%) & 83.69\% (-0.23\%) \\ \midrule
    13) & Ours  & \textbf{86.12\% (-1.06\%)} & \hltred{83.58\% (+1.33\%)} & \hltred{83.63\% (+1.88\%)} & \hltred{85.23\% (+1.26\%)} & \hltred{87.36\% (+1.70\%)} & \hltred{85.10\% (+1.45\%)} \\
    \bottomrule
    \end{tabular}
    }\vspace{1mm}
\resizebox{1.0\linewidth}{!}{
    \begin{tabular}{
    r
    p{4.cm}
    >{\centering\arraybackslash}p{2.9cm}
    >{\centering\arraybackslash}p{2.9cm}
    >{\centering\arraybackslash}p{2.9cm}
    >{\centering\arraybackslash}p{2.9cm}
    >{\centering\arraybackslash}p{2.9cm}
    >{\centering\arraybackslash}p{2.9cm}}
    \multicolumn{8}{c}{\textbf{{\large{(c) Ablation Studies}}}} \\
    \toprule
    1) &  & RE    & LT    & LL    & SF    & LF    & Avg \\
    \midrule
    2) & No mask involved & 2.10 (+16.02\%) & 2.51 (+32.11\%) & 2.48 (+27.84\%) & 3.02 (+72.57\%) & 1.78 (+3.49\%) & 2.38 (+30.77\%) \\
    3) & Mask as attention only & 1.85 (+2.21\%) & 3.37 (+77.37\%) & 2.16 (+11.34\%) & 2.29 (+30.86\%) & 1.75 (+1.74\%) & 2.27 (+24.73\%) \\
    4) & Mask as RANSAC only & 1.85 (+2.21\%) & 2.16 (+13.68\%) & 2.17 (+11.86\%) & 2.04 (+16.57\%) & 2.16 (+25.58\%) & 2.07 (+13.74\%) \\
    \midrule
    5) & w/o. Triple loss & 2.16 (+19.34\%) & 4.15 (+118.42\%) & 3.30 (+70.10\%) & 2.49 (+42.29\%) & 2.09 (+21.51\%) & 2.84 (+56.04\%) \\
    \midrule
    6) & w/o. Feature extractor & 1.89 (+4.42\%) & 2.54 (+33.68\%) & 2.13 (+9.79\%) & 1.80 (+2.86\%) & 1.79 (+4.07\%) & 2.03 (+11.54\%) \\
    \midrule
    7) & VGG~\cite{simonyan2014very}   & 1.91 (+5.52\%) & 2.89 (+52.11\%) & 2.05 (+5.67\%) & 2.14 (+22.29\%) & 1.88 (+9.30\%) & 2.17 (+19.23\%) \\
    8) & ResNet-18~\cite{he2016deep} & 1.84 (+1.66\%) & 2.30 (+21.05\%) & 2.05 (+5.67\%) & 2.28 (+30.29\%) & 1.85 (+7.56\%) & 2.06 (+13.19\%) \\
    9) & ShuffleNet-v2~\cite{zhang2018shufflenet} & 2.05 (+13.26\%) & 2.85 (+50.00\%) & 2.61 (+34.54\%) & 2.72 (+55.43\%) & 1.99 (+15.70\%) & 2.44 (+34.07\%) \\
    \midrule
    10) & Train from scratch & 1.87 (+3.31\%) & 2.00 (+5.26\%) & 1.98 (+2.06\%) & 1.90 (+8.57\%) & 1.77 (+2.91\%) & 1.90 (+4.40\%) \\
    \midrule
    11) & Ours  & \hltred{1.81} & \hltred{1.90} & \hltred{1.94} & \hltred{1.75} & \hltred{1.72} & \hltred{1.82} \\
    \bottomrule
    \end{tabular}
    }
    \vspace{1mm}
    \caption{Quantitative comparison between ours and all other methods including DNN-based (Row 3, 4) and feature-based (Row $5\sim12$) ones, in terms of errors (a) and robustness (b), as well as ablation studies on mask (Rows 2 $\sim$ 4), triplet loss (Row 5), feature extractor (Row 6), backbones (Rows $7\sim9$) and training strategy (Row 10) in (c). For (b), we calculate the inlier percentage when matched points are within 3 pixels. For each scene, we mark the best solution in red. For the scenes ours beats the others, we mark the 2nd best solution in blue.}
    \label{tab:comp-all-methods}
\end{table} 

We also compare our method with some feature-based solutions. Specially, we choose SIFT~\cite{lowe2004distinctive}, ORB~\cite{rublee2011orb}, LIFT~\cite{yi2016lift} and SOSNet~\cite{tian2019sosnet} as the feature descriptors and choose RANSAC~\cite{fischler1981random} and MAGSAC~\cite{barath2019magsac} as the outlier rejection algorithms, obtaining 8 combinations. We show 3 examples in Fig.~\ref{fig:feature_failure}, where (a)(b) show the 8 combinations produce reasonable but low quality results, and (c) shows one that most of them fail thoroughly. Note that the failure cases caused by low texture or low light condition frequently appears in our dataset, and it may lead to unstable results in real applications such as video stabilization or multi-frame image fusion. In comparison, our method is robust against these challenges.


%
%

\subsubsection{Quantitative comparison.}
%
%
%
%
%

We demonstrate the performance of our method by comparing it with all of the other methods quantitatively. The comparison is based on our dataset and the average $l_2$ distances between the warped points and the human-labeled GT points are evaluated as the error metric. We report the errors for each category and the overall averaged error in Table~\ref{tab:comp-all-methods}, where $\mathcal{I}_{3\times3}$ refers to a $3\times3$ identity matrix as a ``no-warping'' homography for reference. As seen, our method outperforms the others for all categories, except for regular (RE) scenes if compared with feature-based methods. This result is reasonable because in RE scenes, rich texture delivers sufficient high quality features so that it is naturally friendly for the feature-based solutions. Even though, our error is only 5.85\% higher than the best solution in this case, i.e. SIFT~\cite{lowe2004distinctive} + MAGSAC~\cite{barath2019magsac}. For the rest scenes, our method consistently beats the others, especially for the low texture (LT) and low light (LL) scenes, where our error is lower than the 2nd best by 25.78\% and 7.62\% respectively. For the scenes containing small (SF) and large (LF) foreground, although the 2nd best method SOSNet~\cite{tian2019sosnet} + MAGSAC~\cite{barath2019magsac} only loses to ours very slightly (0.57\% and 2.82\%), it cannot well handle the LT and LL scenes, where its errors are higher than the 2nd best by 100.78\% and 109.05\% separately.
It is worth noting that the two solutions involving LIFT~\cite{yi2016lift} feature produce rather stable results for all scenes, but their average errors are higher than ours by at least 12.08\%. As for the DNN-based solutions, the supervised method~\cite{detone2016deep} suffers severely from the generalization problem as demonstrated by its errors being higher than us by at least 142.37\% for all scenes, and the unsupervised method~\cite{nguyen2018unsupervised} also apparently fails in the LT scene, causing over 50\% higher error than ours in this case.

\begin{figure}[t]
  \centering
  \includegraphics[width=0.97\linewidth]{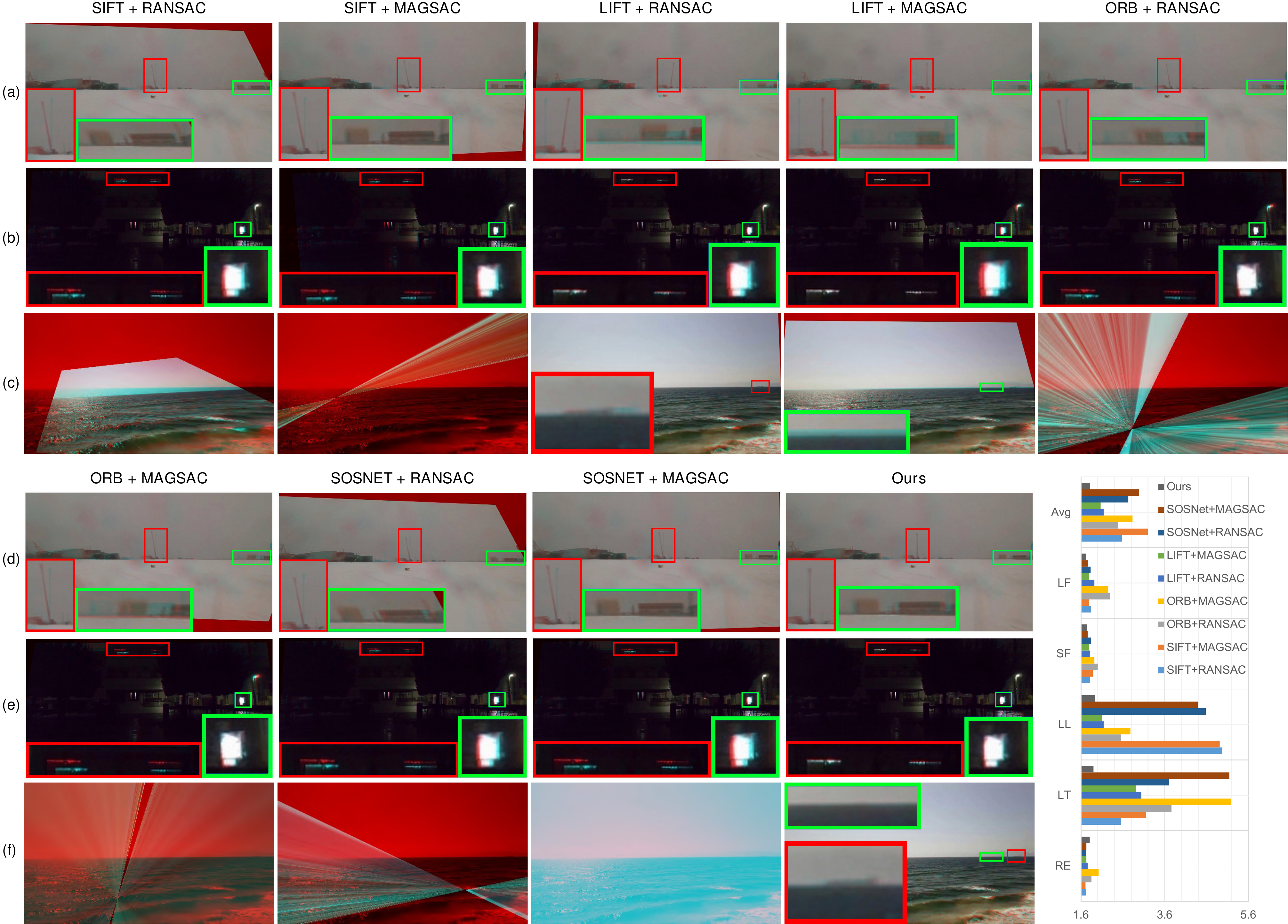}
  \caption{Comparison with 8 feature-based solutions on 3 examples, shown in (a)(d), (b)(e) and (c)(f). For the first 2 examples, our method produces more accurate results, while for the last one but not the least, most of the feature-based solutions fail extremely, which happens frequently for the low texture or low light scenes. We also display the errors by all the methods in bar chart.}
 \label{fig:feature_failure}
\end{figure}

To further evaluate the robustness, a threshold ($3$ pixels) is used to count the percentage of inliners.  Matches that beyond the threshold are considered as outliers. Table~\ref{tab:comp-all-methods}(b) shows the inlier percentage on different scene categories of various methods. As seen, for tough cases, our method achieves the highest robustness compared with other competitors while for regular cases, our performance is on par with the others, which draws similar conclusion as by Table~\ref{tab:comp-all-methods}(a).
Please see Table~\ref{tab:comp-all-methods} and bar charts in Figs.~\ref{fig:sota} and~\ref{fig:feature_failure} for the detailed comparisons.

%
%
%
%
%

\subsection{Ablation Studies}

\subsubsection{Content-aware mask.}\label{subsec:ablation-content-aware-mask}
As mentioned in \secname~\ref{subsec:unsupervised-content-awareness-learning}, the content-aware mask takes effects in two-folds, working as an attention for the feature map, or as a weighting map to reject the outliers.
We verify its effectiveness by evaluating the performance in the case of disabling both or either effect and report the errors in Row 2, 3, 4 of Table~\ref{tab:comp-all-methods}(c).
Specifically, for Row 3 ``Mask as attention only'', Eq.~\ref{eq:l1-warp-Ia-Ib} is modified as $\mathbf{L_n}(I'_a, I_b) = \mathbf{L}(I'_a, I_b) = ||F'_a - F_b||_1$. On the contrary, for Row 4 ``Mask as RANSAC only'', Eq.~\ref{eq:GaGb} is modified as $G_\beta = F_\beta, ~\beta \in \{a,b\}$. As the errors indicate, for most scenes the mask takes effect increasingly by the two roles, except for the scenes LT and LF where disabling one role only may cause the worst result. We also illustrate one example in Row 3, 4 of Fig.~\ref{fig:mask}, where in the case of ``Mask as attention only'' the mask learns to highlight the most attractive edges or texture regions without rejecting the other regions (Column 2). On the contrary, in the case of ``Mask as RANSAC only'', the mask learns to highlight only sparse texture regions (Column 3) as inliers for alignment. In contrast, our method balances the two effects and learn a comprehensive and informative weighting map as shown in Column 4. 

\subsubsection{Feature extractor.}
We also disable the feature extractor to verify its effectiveness, i.e. setting $F_\beta = I_\beta,~\beta \in \{a,b\}$ so that the loss is evaluated on pixel intensity values instead. In this case, the network loses some robustness, especially if applied to images with luminance change, as Fig.~\ref{fig:ablation-feat-ext} shows. As seen, if $f(\cdot)$ is disabled, the masks would be abnormally sparse because the loss reflects only a small falsely ``aligned'' region, causing a wrong homography estimated. In contrast, our results are stable enough thanks to the luminance invariant property of learned features. The errors are listed in Row 6 of Table~\ref{tab:comp-all-methods}(c). 
%
%
%

\subsubsection{Triplet loss.}
We further exam the effectiveness of our triplet loss by removing the term of Eq.~\ref{eq:l1-Ia-Ib} from Eq.~\ref{eq:tripleloss}. As shown in Table~\ref{tab:comp-all-methods}(c) ``w/o. triplet loss'', the triplet loss decreases errors over 50\%, especially beneficial in LT (118.42\% lower error) and LL (70.10\% lower error) scenes, demonstrating that it not only avoids the problem of obtaining trivial solutions, but also facilitates a better optimization.

\subsubsection{Backbone.}
We also exam several popular backbones, including VGG~\cite{simonyan2014very}, ResNet-18~\cite{he2016deep}, and ShuffleNet~\cite{zhang2018shufflenet} for $h(\cdot)$. As seen in Rows $7\sim9$ of Table~\ref{tab:comp-all-methods}(c), the ResNet-18 achieves similar performance as ours (ResNet-34). The VGG backbone is slightly worse than ResNet-18 and ResNet-34. Interestingly, the light-weight backbone ShuffleNet achieves similar performance with other large ones, indicating the potential  application to portable systems of our method. 

%
%
%
%
%
%

\subsubsection{Training strategy.}\label{subsec:ablation-training-strategy}
As aforementioned, we use a two-stage strategy to train the network. To validate this strategy, we conduct an ablation study to train the network from scratch. As Row 10 and 11 of Table~\ref{tab:comp-all-methods}(c) reveal, our training strategy brings a 4.40\% lower error in average, demonstrating its usefulness.



\subsection{Failure Cases}
\begin{figure}[t]
  \centering
  \includegraphics[width=0.24\linewidth]{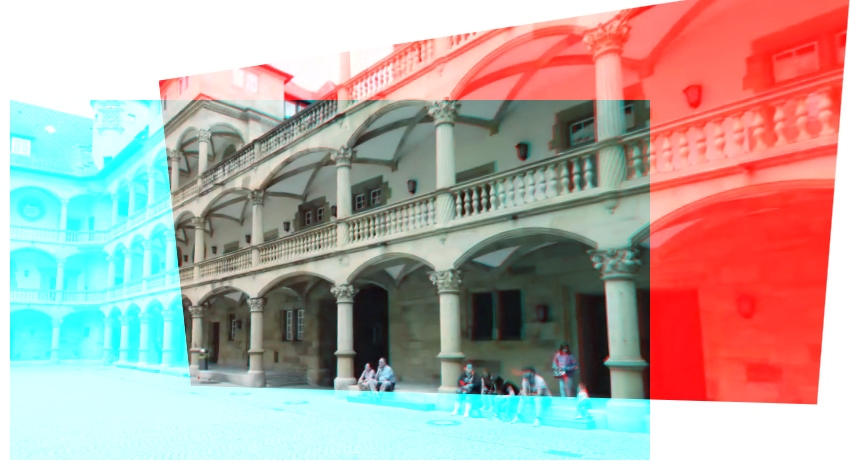}
  \includegraphics[width=0.24\linewidth]{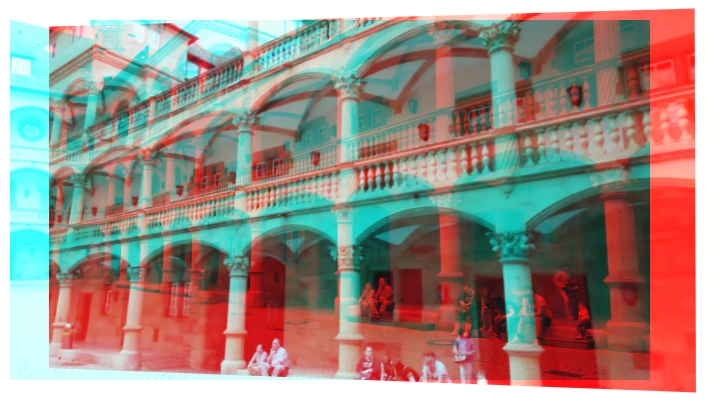}
  \includegraphics[width=0.24\linewidth]{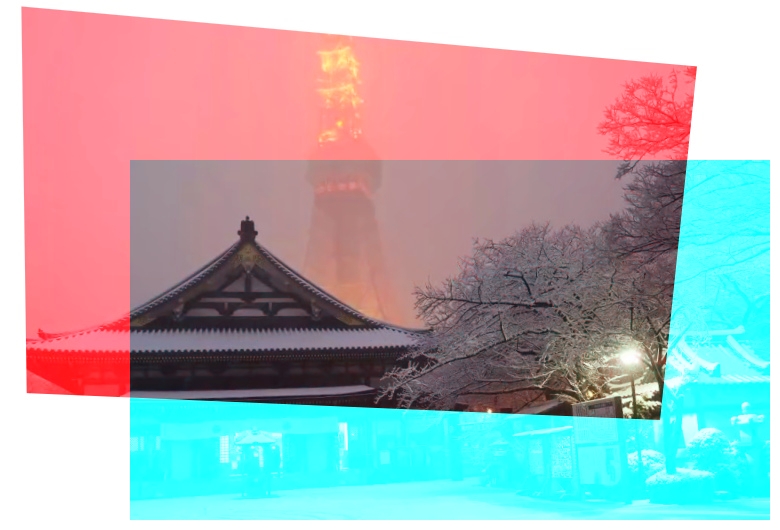}
  \includegraphics[width=0.24\linewidth]{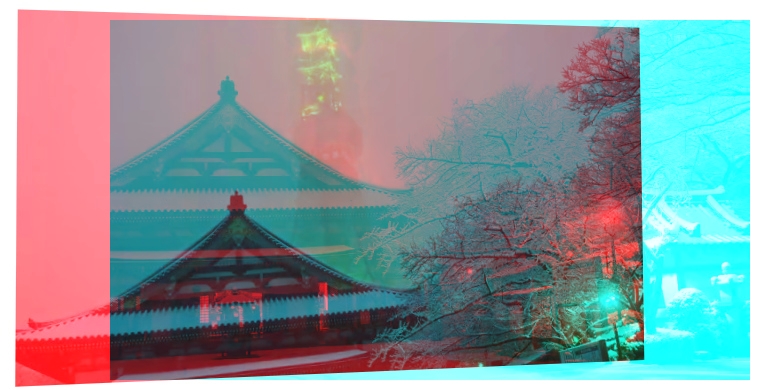}
  \caption{Failure cases. Odd and even columns show the results by SIFT+RANSAC and our method.}\label{fig:failure-case}
\end{figure}

Although our method achieves state-of-the-art performance in small baseline scenes compared with the existing methods, it still has its limitation of being applied to large baseline scenes. The reason behind may lie in the limited perception field of the network which is unable to perceive the alignment information between the two images. With this limitation, our method is unable to be applied to applications relying on large baseline alignment such as image stitching. We show two failure results in Fig.~\ref{fig:failure-case} for large baseline scenes by our method, in comparison with those by SIFT+RANSAC. As seen, SIFT+RANSAC produces stable results for the scenes. We will leave the solution for the large baseline alignment as a future work.

%
%
%

\section{Conclusions}\label{sec:conclu}

We have presented a new architecture for unsupervised deep homography estimation with content-aware capability, for small baseline scenarios. Unlike traditional feature base methods that heavily rely on the quality of image features so as to be vulnerable to low-texture and low-light scenes, or previous DNN-based solutions that pay less attention to the depth disparity issue, our network learns a content-aware mask during the estimation to reject outliers, such that the network can concentrate on the regions that can be aligned by a homography. To achieve it, we have designed a novel triplet loss to enable unsupervised training of our network. Moreover, we present a comprehensive dataset for image alignment. The dataset is divided into 5 categories of scenes, which can be used for the future research of image alignment models, including but not limited to homography, mesh alignment and optical flow. Extensive experiments and ablation studies demonstrate the effectiveness of our network as well as the triplet loss design, and reveal the superiority of our method over the state-of-the-art.



\subsection*{Acknowledgment}
This research was supported in part by National Key Research and Development Program of China under Grant 2017YFA0700800, in part by National Natural Science Foundation of China under Grants (NSFC, No. 61872067 and No. 61720106004) and in part by Research Programs of Science and Technology in Sichuan Province under Grant 2019YFH0016.

\bibliographystyle{splncs04}
\bibliography{egbib}

\begin{thebibliography}{10}
\providecommand{\url}[1]{\texttt{#1}}
\providecommand{\urlprefix}{URL }
\providecommand{\doi}[1]{https://doi.org/#1}

\bibitem{altwaijry2016learning}
Altwaijry, H., Veit, A., Belongie, S.J., Tech, C.: Learning to detect and match
  keypoints with deep architectures. In: {Proc. BMVC} (2016)

\bibitem{baker2004lucas}
Baker, S., Matthews, I.: Lucas-kanade 20 years on: A unifying framework.
  International Journal of Computer Vision  \textbf{56}(3),  221--255 (2004)

\bibitem{barath2019magsac}
Barath, D., Matas, J., Noskova, J.: Magsac: marginalizing sample consensus. In:
  {Proc. CVPR}. pp. 10197--10205 (2019)

\bibitem{bay2006surf}
Bay, H., Tuytelaars, T., Van~Gool, L.: Surf: Speeded up robust features. In:
  {Proc. ECCV}. pp. 404--417 (2006)

\bibitem{bian2017gms}
Bian, J., Lin, W.Y., Matsushita, Y., Yeung, S.K., Nguyen, T.D., Cheng, M.M.:
  Gms: Grid-based motion statistics for fast, ultra-robust feature
  correspondence. In: {Proc. CVPR}. pp. 4181--4190 (2017)

\bibitem{brown2003recognising}
Brown, M., Lowe, D.: Recognising panoramas. In: {Proc. ICCV}. p.~1218 (2003)

\bibitem{detone2016deep}
DeTone, D., Malisiewicz, T., Rabinovich, A.: Deep image homography estimation.
  arXiv preprint arXiv:1606.03798  (2016)

\bibitem{evangelidis2008parametric}
Evangelidis, G.D., Psarakis, E.Z.: Parametric image alignment using enhanced
  correlation coefficient maximization. {IEEE Trans. on Pattern Analysis and
  Machine Intelligence}  \textbf{30}(10),  1858--1865 (2008)

\bibitem{fischler1981random}
Fischler, M.A., Bolles, R.C.: Random sample consensus: a paradigm for model
  fitting with applications to image analysis and automated cartography.
  Communications of the ACM  \textbf{24}(6),  381--395 (1981)

\bibitem{gelfand2010multi}
Gelfand, N., Adams, A., Park, S.H., Pulli, K.: Multi-exposure imaging on mobile
  devices. In: Proc. ACM Multimedia. pp. 823--826 (2010)

\bibitem{godard2019digging}
Godard, C., Mac, O., Firman, M., Brostow, G.J.: Digging into self-supervised
  monocular depth estimation. In: {Proc. ICCV}. pp. 3828--3838 (2019)

\bibitem{guo2016joint}
Guo, H., Liu, S., He, T., Zhu, S., Zeng, B., Gabbouj, M.: Joint video stitching
  and stabilization from moving cameras. {IEEE Trans. on Image Processing}
  \textbf{25}(11),  5491--5503 (2016)

\bibitem{hartley2003multiple}
Hartley, R., Zisserman, A.: Multiple view geometry in computer vision.
  Cambridge university press (2003)

\bibitem{he2016deep}
He, K., Zhang, X., Ren, S., Sun, J.: Deep residual learning for image
  recognition. In: {Proc. CVPR}. pp. 770--778 (2016)

\bibitem{holland1977robust}
Holland, P.W., Welsch, R.E.: Robust regression using iteratively reweighted
  least-squares. Communications in Statistics-theory and Methods
  \textbf{6}(9),  813--827 (1977)

\bibitem{ilg2017flownet}
Ilg, E., Mayer, N., Saikia, T., Keuper, M., Dosovitskiy, A., Brox, T.: Flownet
  2.0: Evolution of optical flow estimation with deep networks. In: {Proc.
  CVPR}. pp. 2462--2470 (2017)

\bibitem{jaderberg2015spatial}
Jaderberg, M., Simonyan, K., Zisserman, A., et~al.: Spatial transformer
  networks. In: Advances in Neural Information Processing Systems. pp.
  2017--2025 (2015)

\bibitem{kingma2014adam}
Kingma, D.P., Ba, J.: Adam: A method for stochastic optimization. arXiv
  preprint arXiv:1412.6980  (2014)

\bibitem{lin2017direct}
Lin, K., Jiang, N., Liu, S., Cheong, L.F., Do, M., Lu, J.: Direct photometric
  alignment by mesh deformation. In: {Proc. CVPR}. pp. 2405--2413 (2017)

\bibitem{liu2016meshflow}
Liu, S., Tan, P., Yuan, L., Sun, J., Zeng, B.: Meshflow: Minimum latency online
  video stabilization. In: {Proc. ECCV}. pp. 800--815 (2016)

\bibitem{liu2013bundled}
Liu, S., Yuan, L., Tan, P., Sun, J.: Bundled camera paths for video
  stabilization. ACM Trans. on Graphics  \textbf{32}(4), ~78 (2013)

\bibitem{liu2014fast}
Liu, Z., Yuan, L., Tang, X., Uyttendaele, M., Sun, J.: Fast burst images
  denoising. ACM Trans. on Graphics  \textbf{33}(6), ~1--9 (2014)

\bibitem{lowe2004distinctive}
Lowe, D.G.: Distinctive image features from scale-invariant keypoints.
  International Journal of Computer Vision  \textbf{60}(2),  91--110 (2004)

\bibitem{lucas1981iterative}
Lucas, B.D., Kanade, T., et~al.: An iterative image registration technique with
  an application to stereo vision. In: {Proc. IJCAI} (1981)

\bibitem{ma2019locality}
Ma, J., Zhao, J., Jiang, J., Zhou, H., Guo, X.: Locality preserving matching.
  International Journal of Computer Vision  \textbf{127}(5),  512--531 (2019)

\bibitem{mur2015orb}
Mur-Artal, R., Montiel, J.M.M., Tardos, J.D.: Orb-slam: a versatile and
  accurate monocular slam system. IEEE Trans. on robotics  \textbf{31}(5),
  1147--1163 (2015)

\bibitem{nguyen2018unsupervised}
Nguyen, T., Chen, S.W., Shivakumar, S.S., Taylor, C.J., Kumar, V.: Unsupervised
  deep homography: A fast and robust homography estimation model. IEEE Robotics
  and Automation Letters  \textbf{3}(3),  2346--2353 (2018)

\bibitem{revaud2016deepmatching}
Revaud, J., Weinzaepfel, P., Harchaoui, Z., Schmid, C.: Deepmatching:
  Hierarchical deformable dense matching. International Journal of Computer
  Vision  \textbf{120}(3),  300--323 (2016)

\bibitem{rublee2011orb}
Rublee, E., Rabaud, V., Konolige, K., Bradski, G.R.: Orb: An efficient
  alternative to sift or surf. In: {Proc. ICCV}. vol.~11, pp. 2564--2571 (2011)

\bibitem{simon2000markerless}
Simon, G., Fitzgibbon, A.W., Zisserman, A.: Markerless tracking using planar
  structures in the scene. In: Proc. International Symposium on Augmented
  Reality. pp. 120--128 (2000)

\bibitem{simonyan2014very}
Simonyan, K., Zisserman, A.: Very deep convolutional networks for large-scale
  image recognition. arXiv preprint arXiv:1409.1556  (2014)

\bibitem{tian2019sosnet}
Tian, Y., Yu, X., Fan, B., Wu, F., Heijnen, H., Balntas, V.: Sosnet: Second
  order similarity regularization for local descriptor learning. In: {Proc.
  CVPR}. pp. 11016--11025 (2019)

\bibitem{weinzaepfel2013deepflow}
Weinzaepfel, P., Revaud, J., Harchaoui, Z., Schmid, C.: Deepflow: Large
  displacement optical flow with deep matching. In: {Proc. CVPR}. pp.
  1385--1392 (2013)

\bibitem{wronski2019handheld}
Wronski, B., Garcia-Dorado, I., Ernst, M., Kelly, D., Krainin, M., Liang, C.K.,
  Levoy, M., Milanfar, P.: Handheld multi-frame super-resolution. ACM Trans. on
  Graphics  \textbf{38}(4),  1--18 (2019)

\bibitem{yi2016lift}
Yi, K.M., Trulls, E., Lepetit, V., Fua, P.: Lift: Learned invariant feature
  transform. In: {Proc. ECCV}. pp. 467--483. Springer (2016)

\bibitem{zaragoza2013projective}
Zaragoza, J., Chin, T.J., Brown, M.S., Suter, D.: As-projective-as-possible
  image stitching with moving dlt. In: {Proc. CVPR}. pp. 2339--2346 (2013)

\bibitem{zhang2014parallax}
Zhang, F., Liu, F.: Parallax-tolerant image stitching. In: {Proc. CVPR}. pp.
  3262--3269 (2014)

\bibitem{zhang2019learning}
Zhang, J., Sun, D., Luo, Z., Yao, A., Zhou, L., Shen, T., Chen, Y., Quan, L.,
  Liao, H.: Learning two-view correspondences and geometry using order-aware
  network. In: {Proc. ICCV}. pp. 5845--5854 (2019)

\bibitem{zhang2018shufflenet}
Zhang, X., Zhou, X., Lin, M., Sun, J.: Shufflenet: An extremely efficient
  convolutional neural network for mobile devices. In: {Proc. CVPR}. pp.
  6848--6856 (2018)

\bibitem{zhang2000flexible}
Zhang, Z.: A flexible new technique for camera calibration. {IEEE Trans. on
  Pattern Analysis and Machine Intelligence}  \textbf{22}(11),  1330--1334
  (2000)

\bibitem{zhou2017unsupervised}
Zhou, T., Brown, M., Snavely, N., Lowe, D.G.: Unsupervised learning of depth
  and ego-motion from video. In: {Proc. CVPR}. pp. 1851--1858 (2017)

\bibitem{zou2012coslam}
Zou, D., Tan, P.: Coslam: Collaborative visual slam in dynamic environments.
  {IEEE Trans. on Pattern Analysis and Machine Intelligence}  \textbf{35}(2),
  354--366 (2012)

\end{thebibliography}
\end{document}